\documentclass[runningheads]{llncs}
\usepackage{graphicx}
\usepackage{caption}
\usepackage{subfig}
\usepackage{verbatim}

\begin{document}
\title{\#MeTooMaastricht: Building a chatbot to assist survivors of sexual harassment}
\titlerunning{\#MeTooMaastricht: Chatbot for assisting survivors of sexual harassment}
%
\author{Tobias Bauer$^\star$ \and Emre Devrim$^\star$ \and Misha Glazunov$^\star$ \and William Lopez Jaramillo$^\star$ \and Balaganesh Mohan\thanks{Denotes equal contribution} \and Gerasimos Spanakis}
\authorrunning{T. Bauer, E. Devrim, M. Glazunov, W. Lopez, B. Mohan, G. Spanakis}
%
\institute{Maastricht University, Department of Data Science and Knowledge Engineering \email{jerry.spanakis@maastrichtuniversity.nl}
}
\maketitle              

\begin{abstract}
Inspired by the recent social movement of \#MeToo, we are building a chatbot to assist survivors of sexual harassment cases (designed for the city of Maastricht but can easily be extended). The motivation behind this work is twofold: properly assist survivors of such events by directing them to appropriate institutions that can offer them help and increase the incident documentation so as to gather more data about harassment cases which are currently under reported. We break down the problem into three data science/machine learning components: harassment type identification (treated as a classification problem), spatio-temporal information extraction (treated as Named Entity Recognition problem) and dialogue with the users (treated as a slot-filling based chatbot). We are able to achieve a success rate of more than 98\% for the identification of a harassment-or-not case and around 80\% for the specific type harassement identification. Locations and dates are identified with more than 90\% accuracy and time occurences prove more challenging with almost 80\%. Finally, initial validation of the chatbot shows great potential for the further development and deployment of such a beneficial for the whole society tool.  

\end{abstract}

\section{Introduction}
As one of the most influential social movements in recent years, \#MeToo has enabled sexual harassment to rise to the surface that usually does not get the attention required \cite{karlekar2018safecity}. There are various types of sexual harassment such as verbal, physical or non-verbal issues in real life and unfortunately, those are some of the most under-reported criminal offenses. Most survivors (we intentionally use the terminology ``survivors" instead of ``victims") may not be willing to go to the police or reveal these issues on social media or even people around, although they are affected mentally or physically or both. There are plenty of reasons for this under reporting, for example, the feeling of shame or embarrassment \cite{binder1981}.

In this nonprofit project, \#MetooMaastricht, we aim to help sexual harassment survivors in the city of Maastricht, Netherlands. Therefore, we introduce the idea of an intelligent tool (namely a chatbot), which can retrieve crucial information from survivors’ texts such as the types of harassment as well as the time and location of the event in order to suggest the best set of actions. 

Bearing in mind the previous studies in sexual harassment and text mining techniques our main research questions are defined as follows:
\begin{itemize}
    \item How can we best design and implement an intelligent chatbot in order to advise people affected by harassment cases?
    \item How can we successfully classify different types of harassment cases based on short texts by using text classification techniques?
    \item Can we extract time and location information from these texts?
    \item How can we use the information extracted  from our models in our final product, a chatbot, for proper guidance to survivors? 
\end{itemize}{}

\section{Related Work}
Most of the work in this project is based on concepts and techniques used in the domain of natural language processing (NLP), so in this section, we set the theoretical framework of our project.

\subsection{Language representation}
Getting from raw text to computer-based language representations is a crucial task in NLP \cite{salton1983introduction}. We briefly describe the most influential ones here: traditional sparse representations (word count vectors etc.) and modern dense representations (word embeddings etc.).

\subsubsection{Sparse representations} The most basic representations of text requires simply counting terms and represent different texts as rows and frequency of each possible term as columns. This approach would result in higher values for more repetitive words and longer texts, advanced techniques to find out relative importance of a term were derived such as TF-IDF vectors \cite{zhang2011comparative}. These vectors consist of two terms; the first one is Term Frequency (TF), which is the ratio of a specific term in a document. The second one is Inverse Document Frequency (IDF) that is equal to the logarithm of the ratio of the total number of documents over the number of documents containing such term within the corpus. Those vectors can be created based on various input types such as words, characters or combination of N terms (N-grams) \cite{cavnar1994n}.

\subsubsection{Word embeddings}
The motivation behind finding better representations for categorical data comes from the limitations of the traditional use of one-hot encoding mapping of categorical variables, where each category is mapped to a high N-dimensional vector consisting of a single ``one" representing a specific value in the variable category, and N-1 zeroes alongside representing the other possible values for the same variable.
To overcome the limitations present in one hot encoding representations, approaches such as Word2vec models have been used in NLP. These models create a dense high dimensional vector representation for each unique word in the corpus of a text input. The vectors obtained are positioned in the vector space such that words that share the same context or are similar are close to one another in that space \cite{mikolov2013efficient}. The two main model architectures used in the Word2vec algorithm are: continuous bag-of-word (CBOW) and skip-gram (SG) models. The main difference between the two of them is that while CBOW takes multiple context of each word as inputs and tries to predict the word corresponding to its context, skip-gram uses the target word to predict the context \cite{rong2014word2vec}.


\subsubsection{Document/paragraph embeddings}
Paragraph or document vector (Doc2vec) is the extended version of Word2vec such that Word2vec learns the d-dimensional representation of words while Doc2vec aims to learn projection of documents into dimensional space. For this purpose, the authors of the Doc2vec simply introduced an additional document vector along with word vectors into Word2vec \cite{le2014distributed}. Therefore, while training the word vectors, the document vector is trained as well, that gives us the numeric representation of the document. 
Similar to Word2vec, Doc2vec has two main models which are Distributed Memory (DM) and Distributed Bag of Words (DBOW). DM is analogous to CBOW that uses document feature vector in addition to surrounding words to predict the target word. On the other hand, DBOW is similar to skip-gram that tries to predict randomly sampled words from the paragraph as outputs.

\subsubsection{State-of-the-art language models and representations}
By combining the latest achievements in language modelling by means of transformers based on self-attention with the idea of deep contextualized word-piece embeddings together with pretraining universal language model, several NLP and AI research groups introduced universal language models that can be subsequently fine-tuned for a specific NLP task.

Google AI group introduced the so-called bidirectional encoder representations from transformers or BERT for short \cite{devlin2018bert}. Google has made BERT code and implementation available, as well as pre trained BERT models on different languages on huge amounts of data where only minor changes can be done to the model to fine tune it to the tasks needed. On top of this research several frameworks have incorporated the current state-of-the-art models such as DeepPavlov \cite{burtsev-etal-2018-deeppavlov}, a python library that builds upon BERT, and many others allowing the user to combine them to improve on many NLP tasks.

\subsection{Text classification}
Text classification is widely used as part of supervised machine learning to tackle similar problems such as sentiment analysis or categorization of articles. Prior to the 1990s, the most common approach were rule-based classification systems, which were manually constructed for each class based on expert opinion \cite{sebastiani2002machine}. Machine Learning techniques have started to dominate old-fashioned rule-based systems in the following decades, as they help to decrease a remarkable amount of engineering effort on rule construction. Text representations (as discussed in the previous paragraphs) play an important role here. Different models can be applied based on the representation basis (TF-IDF vectors, word embeddings, etc.) or the techniques (traditional machine learning algorithms like logistic regression, support vector machines, etc. or deep learning models like recurrent neural networks).

\subsection{Named Entity Recognition}
Named Entity Recognition (NER) is an NLP task that attempts extracting the so called ``named entities" from a text. Named entities may include persons, organizations, locations, time, etc. The most common classical way of NER is based on sequence model tagging like Conditional Random Frields (CRF) \cite{Lafferty:2001:CRF:645530.655813}. State-of-the-art methods of NER are also based on the fine-tuning of pre-trained universal language models (such as BERT which was described previously).

One of the challenges in NER is disambiguation:  tagging a named entity appropriately frequently implies knowledge about the world than cannot be deduced from the formal text analysis only. To that end various knowledge bases and semantic ontologies may be of use. Some of them aim at the specific lexical areas such as WordNet \cite{miller1998wordnet} that allows handling of synonym/antonym words together with a simple hierarchy of hypernyms and hyponyms. Other techniques aim at constructing universal knowledge graphs that represent all the possible knowledge concepts within a single graph with complex and diverse links between them like Wikidata \cite{Vrandecic:2014:WFC:2661061.2629489} which stores information from Wikipedia in a structured way available for online querying.

\subsection{Chatbots}
Chatbot technology was firstly introduced with the implementation of ELIZA in 1964. It was the first program to make Natural language conversation with a computer possible \cite{weizenbaum1976computer}. ELIZA tackled five problems of a chatbot ``the identification of critical words, the discovery of a minimal context, the choice of appropriate transformations, the generation of responses appropriate to the transformation or in the absence of critical words". These are the basic rules still applicable even in modern chatbots.

Today, chatbots have come a long way, and together with more complex NLP modules are used in many business setups for automatic answering and other functions. Some of the most used frameworks include Facebook's wit.ai\footnote[1]{https://wit.ai/} and Google's Dialogflow API\footnote[1]{https://dialogflow.com/}. The operation of modern chatbots does not require any stand-alone platforms; they can be integrated into massively used messaging platforms such as Facebook Messenger, Google Assistant or Telegram.

\section{Methodology}
To answer our research questions, we used data available from SafeCity\footnote[1]{https://safecity.in/} regarding previous harassment reports written by survivors in India. Based on this data, we have trained models with different approaches to classify the cases into different kinds of harassment. Then, by using harassment cases correctly identified by the classifier, we aimed to extract spatio-temporal subject information to properly assist the survivor. This assistance  consists of a set of instructions recommended by the chatbot (our final product). All the inputs and end products of this project are designed for English language.

\subsection{Dataset}
\label{met:dat}
\begin{figure}[h]
\centering
\includegraphics[width=0.7\textwidth]{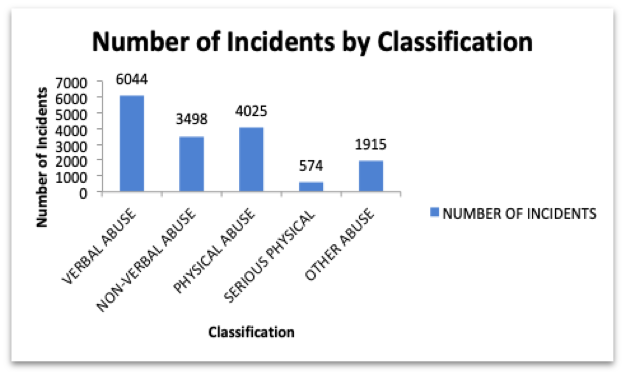}
\caption{Number of harassment types in SafeCity}
\label{haras_types}
\end{figure}

The SafeCity reports contain around 12,000 precise texts in English mainly mentioning  commenting, ogling and groping issues. Moreover, there are more severe physical harassment cases mentioned as well. Also, it should be underlined that a report naturally may include more than one types of harassment. Figure \ref{haras_types} shows the distribution of several types of harassment in such reports used for this project. 

\vspace{-1.1cm}

\subsubsection{Text pre-processing}
 We applied the following pre-processing pipeline by taking into accout the nature of the reports provided by SafeCity. 

\begin{itemize}
    \item \textbf{Contraction handling:} Replacing word contractions such as \textit{I'm} with their unabbreviated form \textit{I am} taking into account misspellings such as \textit{Im}. This was done using regular expressions.
    \item \textbf{Special character removal:} Removing special characters such as \$ and double spaces. This was done using regular expressions as well.
    \item \textbf{Spelling correction:} Simple spelling correction function available in Python was added that uses Levenshtein distance \cite{norvig2007write}.
    \item \textbf{Negation handling:} Simple negation handling approach was used in order to identify the word \textit{not} and finding an antonym for the following word, then replacing both not and its following word with the antonym. This was done using the Wordnet synonym-antonym lexicon from the NLTK in Python \cite{miller1998wordnet} following a similar approach to \cite{krebs2018social}.
    \item \textbf{Lemmatization:} In the feature extraction process for Text Classification models, the corpus was lemmatized in both Bag of Words and Embeddings approaches. This was done using the SpaCy \cite{spacy2}.
    \item \textbf{Lower case:} For the majority of tasks (except Named-entity Recognition) the text was converted to lowercase, since this reduced the corpus size and made no difference in most of the tasks.
    \item \textbf{Part-of-Speech Tags:} We used SpaCy again to find out the most frequent POS tags to visualize our reports (See word clouds in the Appendix). Additionally, we created some models using only these tags but dropped this idea since we couldn't observe performance improvements.
    
\end{itemize}{}

\subsection{Text classification}
\label{met:class}
In this part of our pipeline, the main goal is to determine whether a report is related to a harassment issue. After that, we want to extract more details about the issue, namely types of the harassment or missing information such as time and location in order to suggest proper actions. This would be helpful for our chatbot, in advising appropriate actions to different types of harassment based on the severity of the case such as recommending psychological or medical support.

The initial step is feature engineering where we transform pre-processed text data into feature vectors based on state-of-the-art techniques. We experimented with traditional techniques (like TF-IDF) and with more modern techniques based on embeddings. In particular, we used Doc2vec, a special version of Word2vec for documents/paragraphs \cite{le2014distributed}.  Logistic Regression and Support Vector Machine models were built by using the representations and their performance are discussed in the results under Section \ref{results}. 

\begin{figure}[h]
\centering
\includegraphics[width=0.7\textwidth]{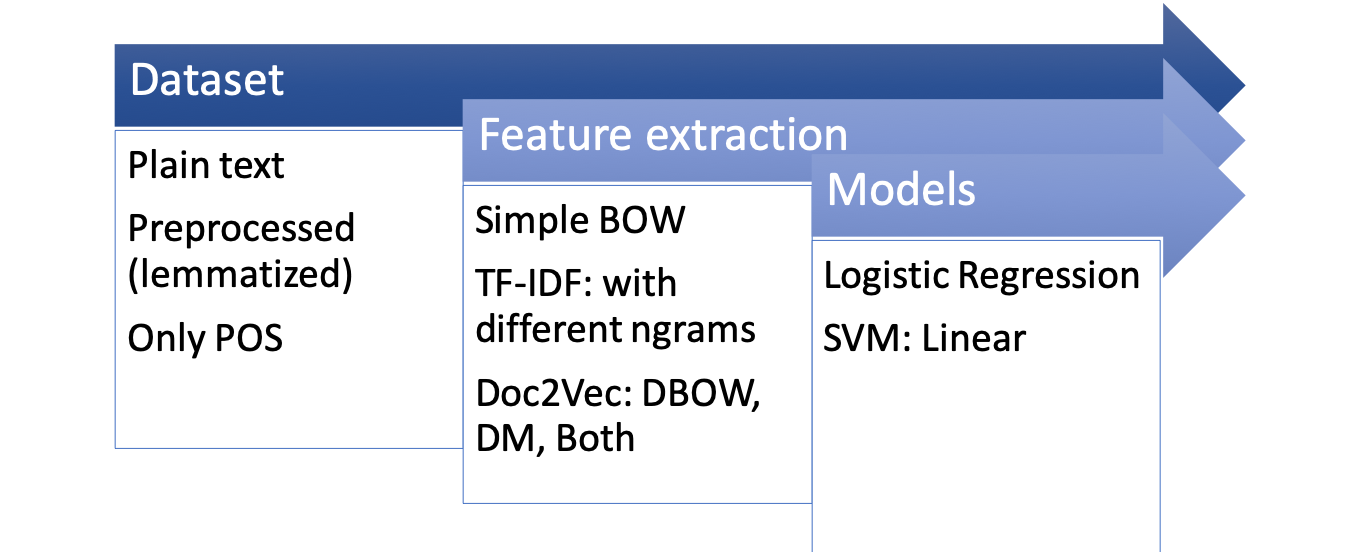}
\caption{Classification flow}
\label{class_flow}
\end{figure}

Figure \ref{class_flow} shows a graphical representation of the workflow used to do the classification task.

\subsection{Named Entity Recognition}
To provide specific assistance to the survivors of harassment, we are interested in the spatial and temporal information of an incident. Spatial information (in this context) is the place where the harassment has occurred. Temporal information, on the other hand, is information about the date and time of the incident. This information can help us provide the right instructions on which actions a survivor should take and can help towards building a spatio-temporal map of harassment cases in Maastricht. To receive these types of information we applied different named entity recognition techniques.

In our project we applied state-of-the-art techniques (mainly based on CRF models) and modern pretraining/finetuning techniques.
For the first part we made use of available solutions in several software packages that are freely distributed, namely the Natural Language Toolkit (NLTK) (Python), the spaCy library (Python) and the Stanford CoreNLP software (Java).

Each of the package exploits different approaches in identifying named entities, so we estimated their capabilities (e.g., entities they are able to extract, the annotation type BIO or BILUO) and identified the one that best fits our needs based on this research study \cite{jiang-etal-2016-evaluating}, namely, Stanford CoreNLP. We considered entities only on the same sentence level, so no co-reference and anaphora resolution have been applied.

\subsubsection{BERT model finetuning}
2018 gave a rise to new successfully applied trends in NLP, namely, unsupervised universal language model pretraining and a subsequent fine-tuning of such a model to the specific NLP task. For this task, we considered Google AI BERT encapsulated into DeepPavlov framework and which was fine-tuned on several widely-used NER datasets for benchmarking such as OntoNotes \cite{Hovy2006} and CoNLL 2003 \cite{TjongKimSang}. Fine-tuning is basically a form of transfer learning: It is applied using a pre-trained generative language model \cite{JeremyHowardFineTuning}, \cite{devlin2018bert}. Large neural networks have been trained on general tasks such as language modelling and then fine-tuned for classification tasks. Particularly, NLP tasks can be fine-tuned with the same single model. In our project we used approach based on universal language model fine-tuning for named entity recognition, namely, pre-trained BERT model for NER task was used for sequence tagging. The framework that is used for Bert NER is DeepPavlov\footnote[1]{http://docs.deeppavlov.ai/en/master/components/ner.html}. The model is based on the Transformer architecture \cite{VaswaniSPUJGKP17}.

\subsubsection{Knowledge base incorporation}
The problem of the wrong NER labeling for cases when a location is labelled as a a person has been  addressed by means of Wikidata. Namely, each person entity is being queried and checked for the presence of the property related to geographical coordinates in the knowledge base. If such property is found there then the person tag is relabelled to the location. 

\subsection{Chatbot}
Chatbots (or more formally conversational agents) have been blooming lately both in research and industry. A chatbot could be defined as a platform that can deal with natural language and queries of the user and respond with appropriate responses. It is important to design an intuitive architecture for conversational user experience.

\subsubsection{Design details}
The conversation flow must be designed in order to gather all the data required to provide correct information. An example of incomplete information follows (``U" stands for user and ``A" stands for the chatbot answer):\\
\par
\textbf{U}: Hello \par
\textbf{A}: Hello, how are you feeling today? \par
\textbf{U}: Not very well. \par
\textbf{A}: May I ask what happened? \par
\textbf{U}: I was walking down the xyz street and a group of men called me mean things. \par
\textbf{A}: I’m so sorry that happened to you. I will try my best to help you with this. \par \medskip

This dialogue doesn't give detailed information such as the type of the incident as well as the exact time and location that are required to provide useful information to the user. ``Mean things" cannot be classified into any kind of legit harassment type. This is why it is important for the chatbot to get direct answers from the user with clear information. To overcome this, we will employ a slot filling based chatbot architecture.

\subsubsection{Slot filling based dialogue modeling}
Slot filling is a way to represent the crucial components that the chatbot should extract from a conversation with any use. In a way, slots are used to represent the semantics of the dialogue. For example, consider the following dialogue (where ``U" stands for the user, ``A" stands for the answer of the chatbot).\\

\textbf{U}: I was walking down in Frankenstraat yesterday evening and a bunch of mean were staring at me! \par
\textbf{P}: I was walking down in \{location\} \{time\} and \{harassment type\}  \par
\textbf{slots:} \begin{itemize}
    \item harasssment type: a bunch of mean were staring at me
    \item location: Frankenstraat
    \item time: yesterday
\end{itemize} \par
\textbf{A}: I am sorry that happened to you! {suggest appropriate action like helpline phone number}\\

Based on the example, we define three slots for our architecture (and we also present some challenges):  \par
\textit{\textbf{@Date}}: Dates can be given in international formats like mm/dd/yy or verbally written like 24th of april etc.. \par
\textit{\textbf{@Time}}: Yesterday cannot be a valid time slot, so the system has to reply with an query for asking for exact time, e.g. I'm sorry that happened to you, I am trying to get the help you need, but I need the exact time frame of the incident.  Alternatively, we can use the system time to understand the meta like yesterday and today. \par
\textit{\textbf{@location}}: Frankenstraat  is a valid slot location. \par \medskip

We also define what will the different intents of the conversation are. In our case, intents are the different type of harassment, and entities are the slots, i.e. date, time and locations. More specifically, we define three intent categories: phsycial abuse, verbal abuse and non-verbal abuse. 

\subsubsection{Approaches for chatbot}
Nowadays, chatbots can be broadly classified as rule-based (scripted) or end-to-end (usually based on deep learning) chatbots. For this project, we experimented with both but decided to proceed with a rule-based approach because of the lack of necessity for a deep learning chatbot and data for training a dataset being very small for  deep learning to be useful.

Telegram\footnote[1]{https://telegram.org/} is a mass communication application used worldwide similar to alternative applications such as Facebook Messenger or WhatsApp\footnote[1]{https://www.whatsapp.com/}. Telegram has support where users can interact with bots by sending them messages, commands and inline requests. The bot created by the API can be specialized for our use case by integrating our NLP platform for question answering. A script was written based on the intents and entities of the several scenarios with appropriate reply vocabularies using python and Telegram API. 

\subsubsection{Chatbot - Dialogue flowchart}
\label{dialogue_flow}
\begin{figure}
\centering
\includegraphics[width=0.95\textwidth]{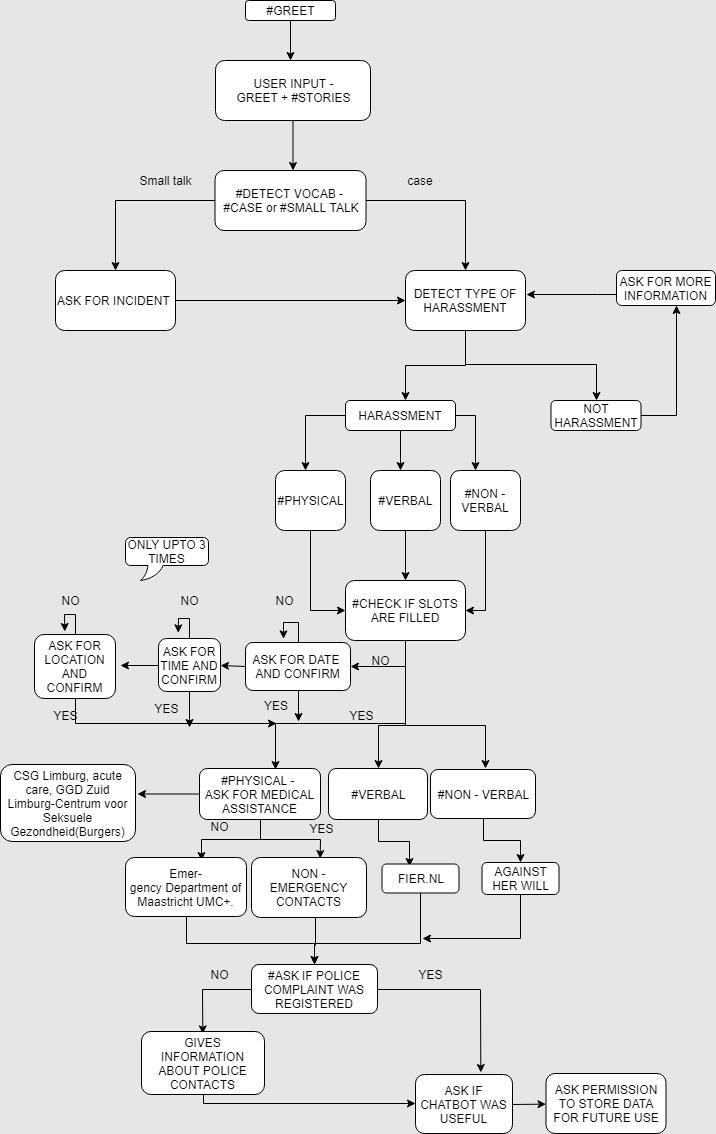}
\caption{Chat dialogue flow}
\label{fig_mt_dnn_arch}
\end{figure}

The ultimate goal of the chatbot is to provide the user with the necessary information based on their input. This has to be as diverse as possible and the conversations must be natural and efficient at the same time. The overall chat workflow is shown in Figure \ref{fig_mt_dnn_arch} and each block will have chatbot reply with unique sentences which were framed with the help of experts \footnote[1]{United Nations University - Maastricht}.

Initially, the chatbot greets the user and asks for information about the possible harassment event. If the user's input is not classified as a harassment case, the chatbot continues to ask. At every step for the user the text sent by the user is concatenated to its previous inputs and it is sent to the classification and named entity recognition system for evaluation. Once the text is classified as harassment, depending on whether the location, date and time information could be retrieved, the chatbot either asks the user for that information if it was missing, or asks the user to confirm the retrieved location, date or time information from the previous input. When there is some slot (location, date or time) missing information the chatbot will ask the user for the details up to 3 times per slot and continue asking for information to fill the next slot. Once all the slots are filled or the attempts to do so have been executed, depending on the type of abuse (physical, verbal, non verbal) identified in the users input the chatbot will provide specific information to the user depending on the case. 

When physical abuse is detected, the chatbot provides information for medical assistance (Emergency Department of Maastricht UMC+), Centrum Seksueel Geweld Limburg (CSG Limburg), Acute care (for crises or emergencies), GGD Zuid Limburg-Centrum voor Seksuele Gezondheid (Burgers).
When verbal abuse is detected, the chatbot provides information of fier.nl \footnote[1]{https://www.fier.nl/chat}, an online chat for support for this kind of abuse. When non-verbal abuse is detected, the chatbot provides information of ``Against her will", another organization specialised in this kind of abuse. Obviously, the specific information provided for each case can be further tailored.

Finally, the user is asked if they have reported the event to the police and relevant information is provided and in the end the chatbot asks the user if they found the process useful and ask for consent to keep the user's data anonymously for further use (e.g. more training data or provide the relevant authorities with more cases).

\section{Results and Validation}
\label{results}
\subsection{Classification models}
We define 4 classification (sub)problems as follows: 
\begin{itemize}
    \item \textbf{Harassment or not:} First of all, we wanted to see that at what level we can diversify a harassment case from any similar short text which is written by a user on the Internet. Therefore, we collected datasets consisting of some user reviews on IMDB,  Amazon or tweets on Twitter as the negative class of our target.
    \item \textbf{Labeling verbal abuses among all harassment reports:} As a next step, we created models in order to catch verbal abuses among all harassment cases. We already had those labels thanks to the SafeCity dataset. 
    \item \textbf{Labeling non-verbal abuses among all harassment reports:} Similar to verbal models. 
    \item \textbf{Labeling physical abuses among all harassment reports:} Since the number of serious physical abuses was low, they were merged to physical abuses. 
\end{itemize}{}

For these models, different datasets were created in which numbers of positive and negative classes were in balance. In order to compare candidate models properly, 30\% of the data were selected as a test set which was stratified by the target. Then, combinations of various text types, feature extraction methods and modeling techniques were implemented as can be seen in Figure \ref{class_flow}. 

In the final models, which are input for the chatbot, two models for each classification problem were created by using pre-processed (lemmatized) text. Those use TF-IDF with up to 3 n-grams and Doc2Vec with Distributed Bag of Words (DBOW) approaches respectively. We decided to use these different approaches since both resulted in a good performance in the test set and ensembling them in the chatbot would give us more robust outcomes.

\begin{figure}[h]
\subfloat[TF-IDF (up to 3 n-grams)]{\includegraphics[width=0.49\textwidth]{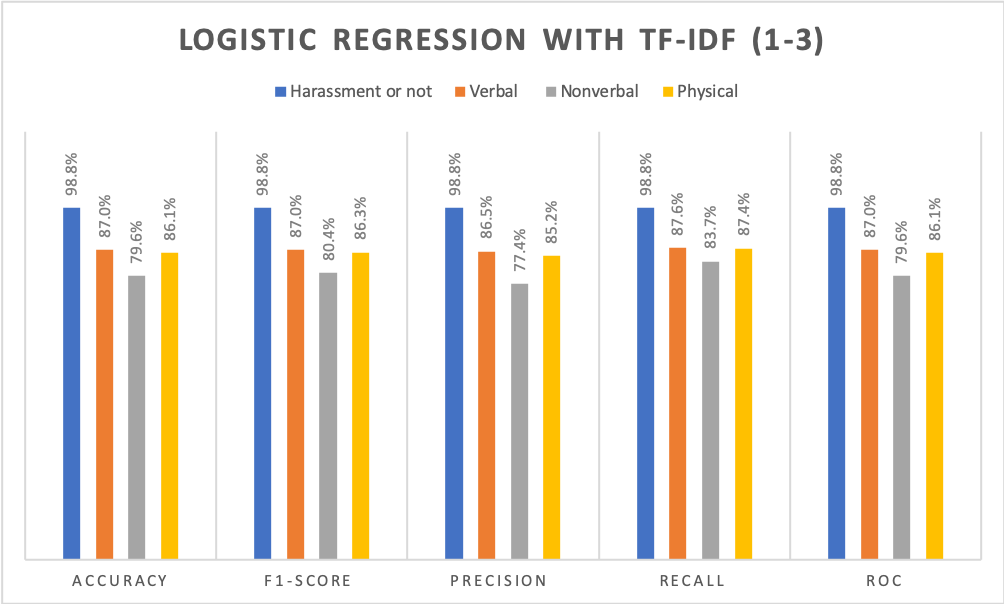}}
\subfloat[Doc2Vec]{\includegraphics[width=0.52\textwidth]{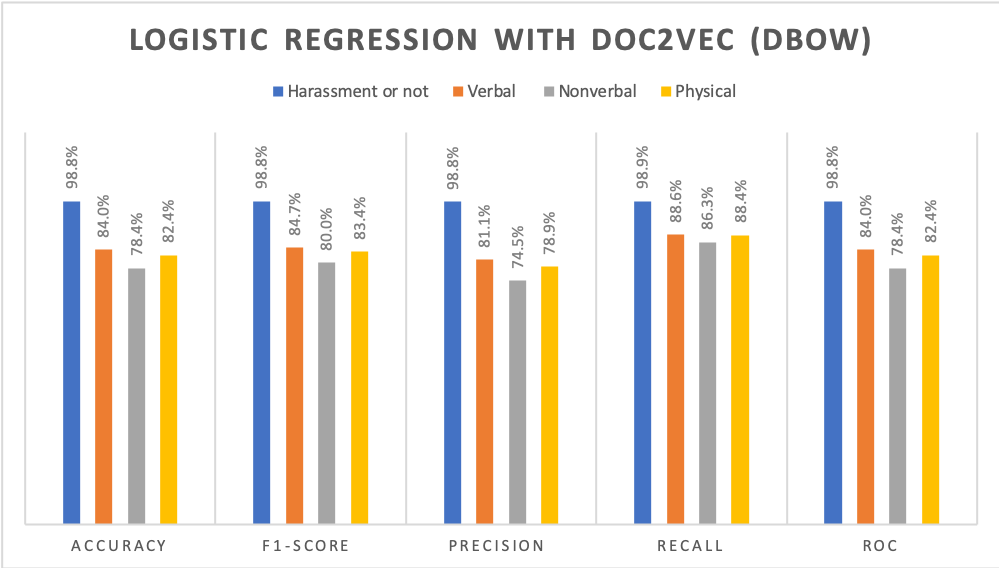}}
\caption{Logistic Regression Models Final Performance}
\label{classresults}
\end{figure}

As the classification model, both use Logistic Regression since it has performed better than SVM and returns the probability that gives us the flexibility to change the cutoff. The chatbot is capable of processing incoming texts through the same steps and classify them. Figure \ref{classresults} shows the performance of final models on test sets.

\vspace{-1.0cm}

\subsection{NER Validation}
For the named entity recognition we did a validation using a self made dataset. We created this dataset by writing 5 short reports of harassment cases. In these reports we set the named entities with placeholder variables. To show that the NER model works for a variety of different named entities we downloaded a list of 12900 city names from around the world from SimpleMaps\footnote[1]{https://simplemaps.com/data/world-cities}. To further verify that the model is able to identify date and time information in a text we chose different formats to represent those information that can be inserted into the reports. Examples for the date format are: ``yesterday", ``5 months ago" or ``on the 5th July 2019". On the other hand examples for a time format are: ``around 10am", ``at 10 o'clock" or ``at night". 

In the next step we inserted these location, date and time information into the reports at the designated positions randomly. Subsequently we put the resulting reports into the different NER models and compared the results provided by these with the original named entities. To avoid cases in which the detected named entities match except for the prefix we removed prefixes from both strings.

\begin{table}[h]
    \centering
    \begin{tabular}{l|c|c|c}
        Classifier & Location & Date & Time \\ \hline
        BERT with Ontonotes & 0.92 & 0.934 & 0.798 \\
        BERT with CoNLL & 0.976 & - & - \\
        Stanford & 0.45 & 0.2 & 0.1 \\
    \end{tabular}
    \caption{Validation results for accuracy}
    \label{tab:ner_results}
\end{table}

\vspace{-1.0cm}

To receive comparable results for the three different used NER models namely Stanford, BERT trained on CoNLL corpora and BERT trained on Ontonotes we generated for each report template 100 variations with randomly picked named entities and used them as input for the models. Table \ref{tab:ner_results} shows the result of these tests. It can be seen that both BERT models deliver reasonable results for the identification of location entities. However the BERT model trained on CoNLL corpora is not able to identify any information about the date or time. However the results produced by BERT are significantly better than the results from the Stanford NER model. The drop of accuracy for time information in the BERT model can be explained by looking at the returned values. Apparently there is some confusion between date and time information.

\subsection{Chatbot Validation}
Because of the complexity of the chatbot dialogue flow we were not able to validate the chatbot entirely. However, we were able to write scripts of specific showcases and compare the responses given by the chatbot with the responses we expected. 

In the first scenario we don't greet the chatbot at all and just report to it an incidence that is clearly a form of physical harassment. We also provide all necessary information about the location, date and time of the incident directly in the first message. Thus the bot just asks us to confirm this information. In the next step we expect that the bot asks if we need medical assistance. We decline that and the bot gives us the contact details of CSG Limburg, acute care and the GGD Zuid Limburg-Centrum voor Seksuele Gezondheid. Afterwards the bot asks us if we reported the incidence to the police. We answer with yes, so the bot does not give us any additional information and just asks us if it was helpful. To try out if everything is working we answer with no. In the last step the bot asks if it can store the data anonymously. We accept this and the bot ends the conversation as expected. 

In the second scenario we greet the bot and introduce ourselves as John in the first message. Thus we expect the bot to ask us about the incident. So the second message we send describes an incident that can be categorized as a form of verbal abuse. But this time we do not provide any information about the location, date or time at all. So we expect the bot to ask us about the location this incident took place. So we tell the bot that this took place ``in Maastricht" and confirm with yes after the it asks us if this is correct. In the next step the bot asks us about the date on which the abuse occurred. Again we give it the answer straight away by replying with ``yesterday" and confirm with yes. Lastly the bot asks us at which time it occurred and we answer with ``at 10am" and confirm once again. In the next step, since the report clearly described an incident of verbal abuse the bot gives us the contact information of fier.nl and asks us if the police was already informed. We reply with ``no" and receive the contact information of the local police department. Afterwards the bot asks us again if it was helpful. We answer with yes this time and the bot then asks for permission to store our data. This time we refuse and the bot says us goodbye and ends the conversation.

In the last scenario we send the bot a message that clearly has nothing to do with any form of sexual harassment. Hence we expect the bot to ask for more information. So in the next message we report an incidence that falls under the category of non-verbal abuse. But again we do not provide any information about the location, date or time. Thus the bot asks us where and when this took place. We reply three times with a message that clearly does not contain any information about the location, date or time. Thus the bot continues by giving us information about ``Against her will" and asking us if it was reported to the police, if the bot was helpful and if it can store the data. 

\begin{figure}
    \centering
    \includegraphics[width=\textwidth]{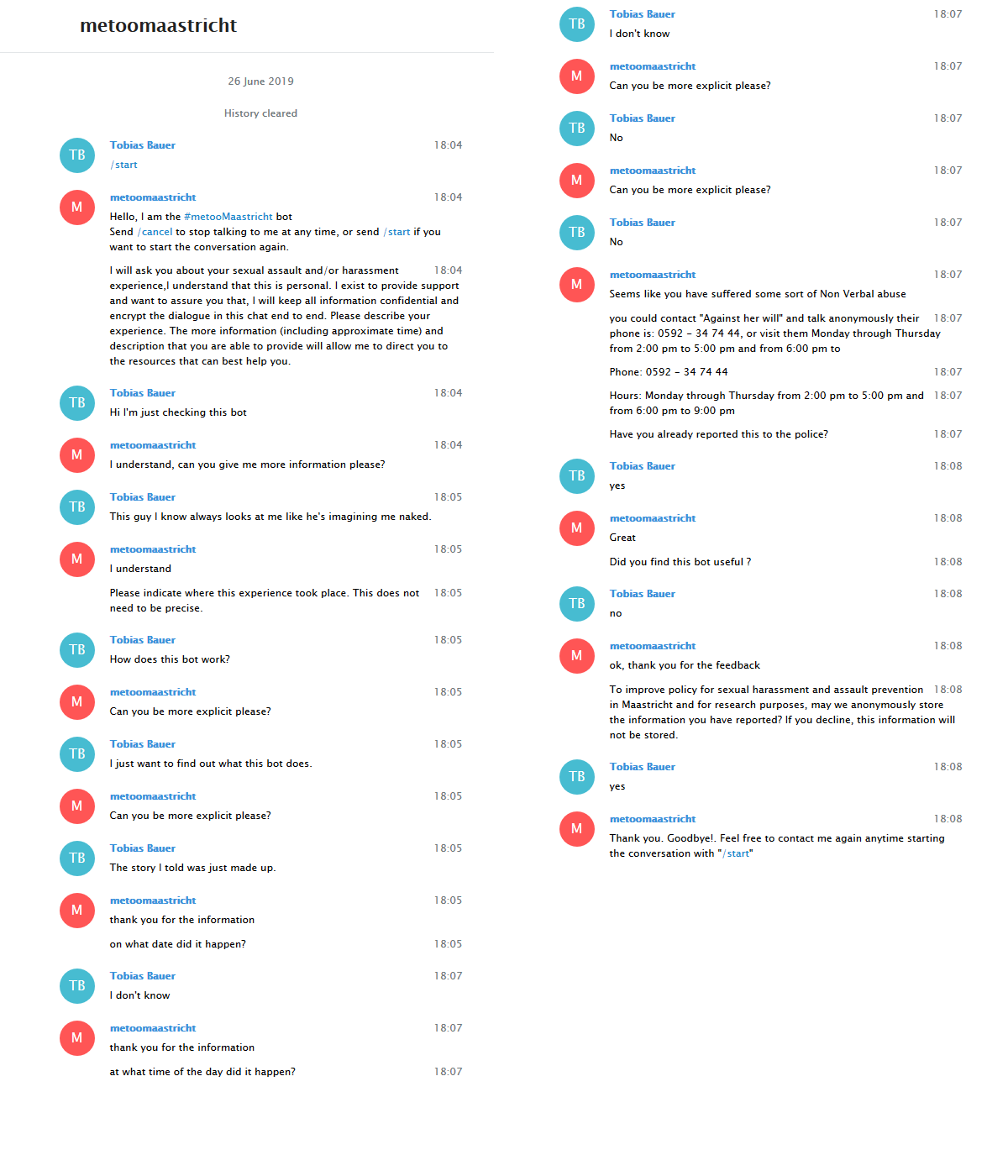}
    \caption{Transcript of scenario 3}
    \label{fig:case3}
\end{figure}

The complete transcripts of the conversations can be found in the Appendix in Figures \ref{fig:case1}, \ref{fig:case2} and \ref{fig:case3}. The responses of the bot match the chat dialogue flow described in section \ref{dialogue_flow}. 

\section{Conclusion}
\#MeToo is a social movement that has attracted great media attention in recent years, especially in social networks. As global awareness is rising, the goal of this work, namely \#MeTooMaastricht is to provide survivors of sexual harassment a safe platform to share their experiences and get proper assistance. To this twofold purpose, we implemented a chatbot using the Telegram API. In order to provide the most appropriate help, we have taken into account various factors related to the incident, such as the type of harassment that was experienced as well as the location, date and time of the incident. The latter proved to be challenging, as it is not trivial to extract accurate spatio-temporal info from a chat text about sexual harassment. 

Classification of the harassment type was successful by reusing data from SafeCity and combining two models: a TF-IDF with up to 3 n-grams and Doc2Vec with Distributed Bag of Words (DBOW) with a Logistic Regression classifier. Results gave an over 80\% accuracy for identification of harassment type. Named entity recognition (NER) was implemented by finetuning BERT state-of-the-art model enhanced by the Wikidata knowledge base and delivered very accurate results for location and dates (90\%) and very satisfactory results for time events (80\%). Finally, a slot-filling based chatbot was implemented so as to encapsulate the classification and NER frameworks into the dialogue flow.

Initial results of this work are really encouraging into ways that survivors of harassment can be assisted by means of data science. However, there are many possible directions for improvement in the future. First of all, the interaction with the chatbot can be improved in terms of what type of language is used. On this end, we plan to further work with social scientists that can run specific focus groups on validating the script flow. Moreover, we want to explore more possibilities on the technical side (e.g. use location or map info so as to enhance the results of NER) and on the security front (e.g. guarantee anonymity and malicious use). Finally, one of our overarching goals is to have a chatbot which is adaptable to each case (e.g. show empathy when needed) and be less ``linear'' in its functionality (e.g. act more freely but still within the script). 

\section{Acknowledgements}
We would like to thank the Safecity website for providing the dataset needed for the first part of this research. We also gratefully acknowledge the valuable contributions of Mary Kaltenberg, post-doctoral fellow at Brandeis University, in building the dialogue flow of the chatbot implemented in this paper.

\bibliographystyle{unsrt}
\bibliography{ref.bib}

\newpage
\onecolumn

\section*{Appendix}

\begin{figure}
    \centering
    \includegraphics[width=\textwidth]{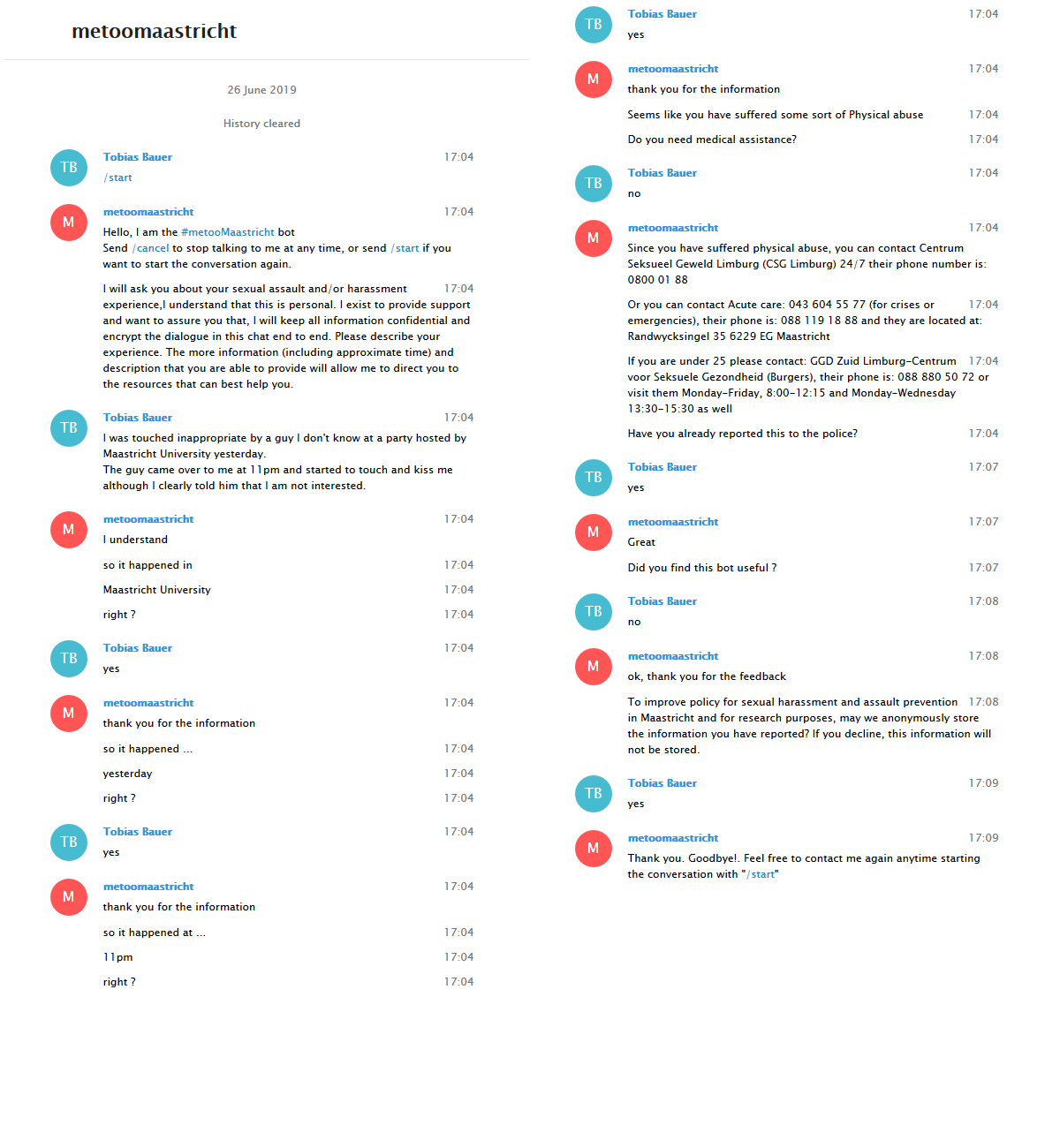}
    \caption{Transcript of scenario 1}
    \label{fig:case1}
\end{figure}

\begin{figure}
    \centering
    \includegraphics[width=\textwidth]{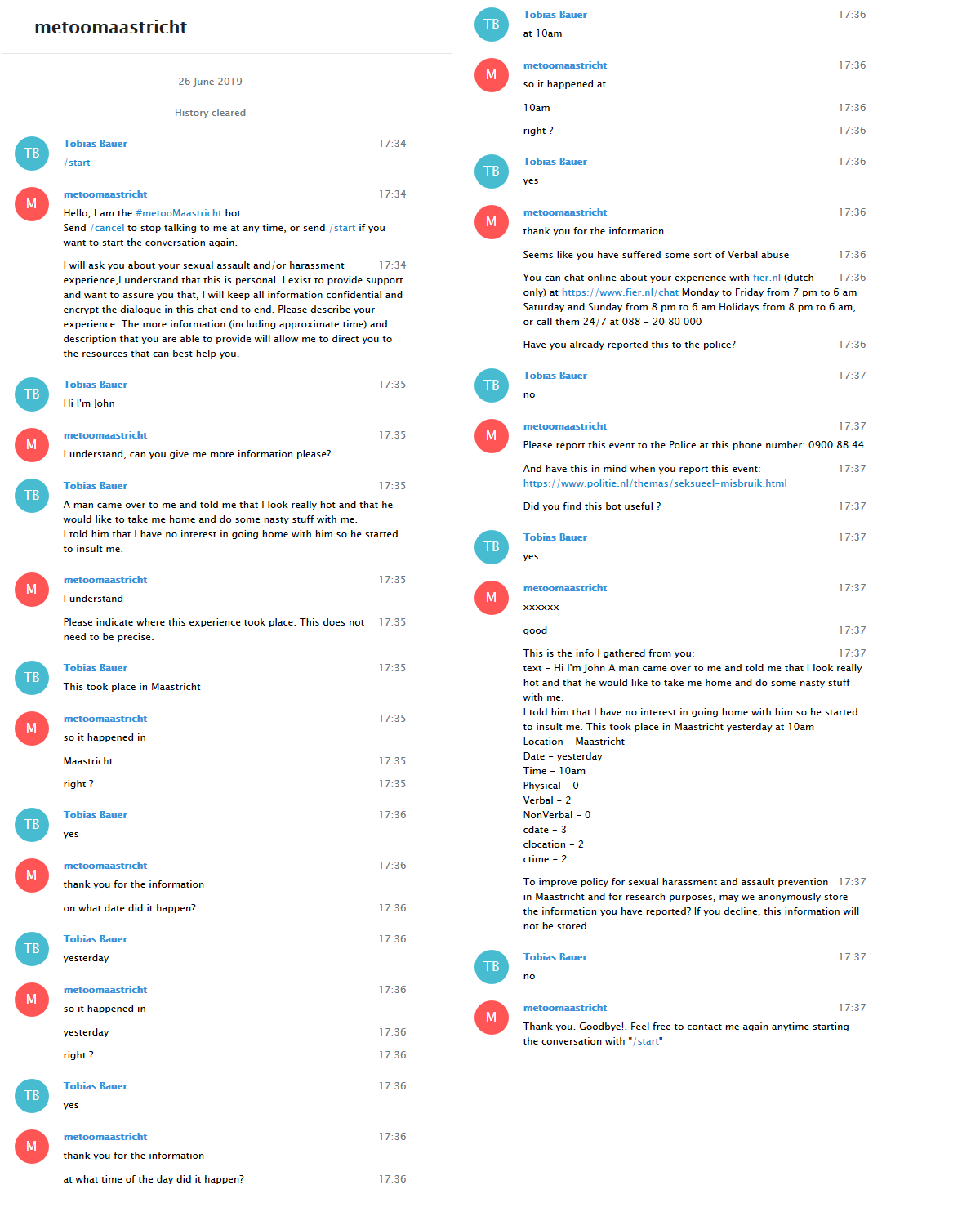}
    \caption{Transcript of scenario 2}
    \label{fig:case2}
\end{figure}

\end{document}